\documentclass[10pt,conference]{IEEEtran}

\usepackage[utf8]{inputenc}
\usepackage[T1]{fontenc} 
\usepackage{amsmath}
\usepackage{graphicx}
\usepackage[cmintegrals]{newtxmath}
\usepackage{bm} 
\usepackage{cite}
\usepackage{booktabs}
\usepackage{array}
\usepackage{url}
\newcolumntype{L}[1]{>{\raggedright\let\newline\\\arraybackslash\hspace{0pt}}m{#1}}

\title{A Graph Neural Network Approach for Scalable and Dynamic IP Similarity in Enterprise Networks}
\author{\IEEEauthorblockN{Hazem M. Soliman, Geoff Salmon, Dusan Sovilij, Mohan Rao}
\IEEEauthorblockA{\textit{RANK Software Inc.} \\
Toronto, Canada \\
\{hazem.soliman, geoff.salmon, dusan.sovilij, mohan.rao\}@ranksoftwareinc.com}}
\date{August 2020}

\begin{document}


\maketitle

\begin{abstract}

Measuring similarity between IP addresses is an important task in the daily operations of any enterprise network. Applications that depend on an IP similarity measure include measuring correlation between security alerts, building baselines for behavioral modelling, debugging network failures and tracking persistent attacks. However, IPs do not have a natural similarity measure by definition. Deep Learning architectures are a promising solution here since they are able to learn numerical representations for IPs directly from data, allowing various distance measures to be applied on the calculated representations. Current works have utilized Natural Language Processing (NLP) techniques for learning IP embeddings. However, these approaches have no proper way to handle out-of-vocabulary (OOV) IPs not seen during training. In this paper, we propose a novel approach for IP embedding using an adapted graph neural network (GNN) architecture. This approach has the advantages of working on the raw data, scalability and, most importantly, induction, i.e. the ability to measure similarity between previously unseen IPs. Using data from an enterprise network, our approach is able to identify similarities between local DNS servers and root DNS servers even though some of these machines are never encountered during the training phase.
\end{abstract}

\section{Introduction}

The problem of measuring how similar a group of machines, denoted with their IPs, are is an important aspect in a large number of network operations and security analysis tasks. In the security context, malicious behavior such as an Advanced Persistent Attack (APT) typically affects multiple machines at a time \cite{Ghafir2014} -- in the initial phases when the adversary is looking for exploits in the network assets, and then after successfully infiltrating the network when the adversary is spreading internally. Alert correlation is of crucial importance in detecting APTs, where individual alerts are not necessarily an indication of an actual attack, but a correlated group of such alerts following a proper attack plan is a significantly better indication of malicious behavior \cite{cao2015preemptive}. Since an alert is typically defined through its attributes, such as IP addresses, port number, ... etc, the problem of alert similarity entails the one of IP similarity, where two alerts are more similar if the IPs involved in both alerts are also similar. The problem under study here, IP similarity, is then a subset of the larger problem of general alert similarity.


Once a security alert is triggered, it is presented to the analyst for further investigation, following which the analyst may choose to tag the alert as false-positive. The problem of the high rate of false-positives in security analytics and behavioral modelling is a well-documented one \cite{sommer2010outside}. One approach here is to ask the analyst for feedback on a small set of alerts, and use this feedback to better label the rest of the larger group of alerts through alert similarity, forming an active-learning loop \cite{veeramachaneni2016ai}. This again leads to the question of alert similarity, and consequently to that of IP similarity.

User Entity Behavior Analytics (UEBA) \cite{shashanka2016user} and Network behavior anomaly detection (NBAD) leverage statistical approaches for threat detection in modern enterprise networks. The most common approach here is to build behavior baselines from historical traffic data and evaluate new incoming data against these baselines. An alert is then triggered if the new traffic deviates significantly from the established baselines. However, this approach suffers greatly with new machines and sparse traffic. For example, consider a load-balanced system where one server is capable of handling most of the requests while a backup server is idle during the majority of time. A UEBA system may trigger an excessive number of alerts for this idle server when the load increases and it has to handle more traffic. One solution is to use not only the data for the machine under study but also the data from similar machines, in order to build a more robust baseline. In this case, the question of similarity between machines is of significant importance.


\subsection{Related Work}
\label{sec:review}

The problem of distance measures between IP addresses has been extensively studied in the literature, particularly in security applications. One of the earliest works is \cite{julisch2003clustering}, where the hierarchy of IP addresses is used to measure their pairwise distances depending on how far they are in the hierarchy tree. This approach has been extended in \cite{coull2011measuring}. Both approaches however have the drawback of being manually designed and not granular enough beyond the main divisions in the tree, such as public versus private IPs.

Another line of work uses the bit representation of IPs and measures distances based on the length of the longest common prefix \cite{shittu2016mining}. This approach has been utilized to group alerts together into an attack graph if the distances between them is below a certain threshold. However, this approach suffers from putting too much faith in IP allocation, where successive IPs are assumed to be assigned  to the most similar machines. With the migration to the cloud and the dynamic scaling of enterprise networks, these assumption are becoming out-dated. Moreover, this approach is not designed to handle similarity between public IPs.

Driven by the limitations of static, manually-written distance measures, and inspired by recent advances in NLP and word embeddings, IP2vec \cite{ring2017ip2vec} was proposed as an extension of the popular word2vec \cite{goldberg2014word2vec} paradigm to the networking domain. The idea behind the approach is to treat network logs as the equivalent of a text corpus in NLP, and entries in the log message, such as IPs and port numbers, as the equivalent of words. Finally, train a neural network on this data and use its hidden layer as the embedding of the network entities of interest. These entities include IPs, ports and network protocols. This approach does not generate similarity per se, instead it generates a numerical representation for each network entity. Various distance functions can be applied on the given numerical representations to get the required similarity. This approach has the advantage of learning representation directly from data without the need for manually designed measures, rendering it more practical and flexible. Recently, this approach has been applied to the detection of persistent attacks in enterprise networks with promising results \cite{burr2020detection}.

\subsection{Proposed Methodology}
\label{sec:method}

In this paper, we take this approach one step further by leveraging graph neural networks (GNNs) to generate an embedding for the nodes of the network communication graph, where each node is a distinct IP. Our motivation behind this approach is as follows:
\begin{itemize}
    \item Network data represents a graph, in particular the flow/connection logs describe the communication graph between the different entities in the network. This makes GNNs a natural choice to get a representation of the graph and its nodes.
    \item Out-of-vocabulary (OOV) problem: this is a well-known problem in NLP concerned with how to handle new words during testing which were not seen during training \cite{bazzi2002modelling}. This problem is more challenging in our case since networking data is much more dynamic than text data. New IPs show up all the time, whether due to new machines joining the network, new external servers that users access, or dynamically allocating IPs in the network. The ability to generate a representation of previously unseen IPs is a crucial feature in a network embedding model. NLP-based approaches fall short here, while a large  training text corpus is usually enough to see almost all relevant words, this is not the case with networks data due to the large number of IPs ($\sim2^{32}$ for IPv4 and $\sim2^{128}$ for IPv6). The same large number of IPs hinders the feasibility of building a model that can handle all possible values. 
    
    The NLP field has some approaches to handle the OOV issue, none of which are applicable to our problem. One common solution is to generate character embeddings, single or few characters, and build a word embedding from the embeddings of its composing characters \cite{kim2016character}. However, IP addresses do not have the same compositional nature as words. While a single character is unlikely to change a word's meaning, two IPs with only one different bit can belong to two machines with completely different jobs. Another approach to address OOV is to have a singular representation for all unseen words. Clearly, this solution is not applicable here. While in NLP tasks the size of OOV is small, in networking due to the huge number of IPs, the number of IPs encountered in any realistic training set is much smaller than the total number of IPs in the internet, and the size of OOV is actually quite significant.
\end{itemize}

The rest of the paper is organized as follows: Section \ref{sec:overview} provides an overview of graph neural networks. Both the system and neural network architectures are discussed in Section \ref{sec:model}. In Section \ref{sec:results} we show the results from a real network dataset. The paper is concluded in Section \ref{sec:conclusion}.

\section{Overview of Graph Neural Networks}
\label{sec:overview}

Graph neural networks comprise a special class of neural networks designed to handle data in a non-Euclidean vector space in the form of a graph \cite{dwivedi2020benchmarking}\cite{wu2020comprehensive}. The goal is to leverage the benefits of deep learning methods, such as distributed representations and end-to-end learning \cite{lecun2015deep}, for application domains in which the data does not fit into the more popular paradigms of convolutional or recurrent neural networks (CNNs, RNNs). Examples of this non-Euclidean data are chemical molecules, citation networks, user-product interaction in e-commerce, and, in our case, communication networks.

One of more popular problems studied in the GNN is known as network embedding, where the goal is to generate low-dimensional vector representation for the graph nodes in line with the graph topology and the node information content \cite{hamilton2017representation}. These representations, also referred to as embeddings, are then used for various tasks such as node classification, clustering, and in our case, measuring similarity between nodes using their vector representation. 

GNNs can be classified into two broad families \cite{dwivedi2020benchmarking}: message-passing convolutional GNNs and Weisfeiler-Lehman GNNs. Message-passing GNNs update a node representation using the representations of its neighbors, through variations of the basic formula:
\begin{equation}
    h_i^{l+1} = f \left(W_1^l h_i^l + \sum_{j \in \mathcal{N}_i} W_2^l h_j^l\right)
\end{equation}

where $h_i^l$ is the representation of node $i$ at layer $l$, $f(.)$ is a non-linear mapping such as ReLU, $W$'s are learnable weight tensors and $\mathcal{N}_i$ is the set of neighbors for node $i$.

One of the major advantages of message-passing GNNs is their inductive nature. A GNN is called inductive if, during the test phase, it can generate a representation for a node it has not seen before during the training phase. This is mainly due to the parameter sharing between all nodes and the message passing operation itself \cite{hamilton2017inductive}. This inductive nature is a key feature that we believe gives GNN the edge over NLP-like approaches for the task of IP embedding.

In summary, our design goals for GNNs in the embedding of network IP addresses are:
\begin{itemize}
    \item Inductive vs transductive: inductive approaches can embed unseen nodes, a feature we need for constantly changing networks.
    \item Edge attributes: compared to other application domains, edge information such as protocol, byte count, ... etc  are particularly important for our use-case.
    \item Anomaly detection: an auto-encoder based loss function is useful here as the GNN would not only provide the embedding but also an anomaly signal if the network instance is very different from the training data.
\end{itemize}


\section{System Description}
\label{sec:model}

\subsection{Architecture}
\label{ssec:arch}

The input data to our system is the set of network logs from the Bro/Zeek monitoring tool\footnote{\url{https://zeek.org/}}. These logs are pre-processed for any data enrichment/cleaning and then used to build the communication graph of the network. An offline training step uses a large historic dataset to train the GNN and save the trained model. During real-time operation, a new batch of data is applied to the model every fixed interval, 10 minutes for example, and the inference operation of the GNN provides the embeddings for the IPs seen during the interval as well as an anomaly signal if needed. The offline training operation can be periodically repeated as deemed necessary. Note that due to the inductive nature of the model, we can do the training on one network and deploy the model on as many networks as desired, resulting in more deployment flexibility.

\subsection{Mathematical Model}
\label{ssec:mathsmodel}

Our model is an extension of the Residual Gated Graph Convolution model proposed in \cite{bresson2017residual}, and shown in \cite{dwivedi2020benchmarking} to be one of the best performing GNN models on a variety of tasks.

\subsubsection{Data Format}
\label{sssec: dataFormat}
The input data to our system comes from Bro/Zeek conn.log files. Table \ref{tab:fields} has a list of the fields we use in our experiments.

\begin{table}[hbt]
  \begin{center}
    \caption{Fields used from Zeek conn.log file}
    \label{tab:fields}
    \begin{tabular}{L{3cm} | L{4cm}}
      \toprule 
      \textbf{Group} & \textbf{Fields}\\
      \toprule 
      IP information & sourceIP, destinationIP \\
      \midrule 
      Port Information & sourcePort, destinationPort \\
      \midrule 
      Protocol Information & protocolService \\
      \midrule 
      Numerical Traffic Features & responseBytes, requestBytes, duration, bytes, responsePackets, requestPackets, 
    responseIPBytes, requestIPBytes \\
      \bottomrule 
    \end{tabular}
  \end{center}
\end{table}

Due to the incredibly high rate of new connections in any real world network, we group connections together into a single graph for a fixed interval, 10 minutes for example. In particular, the data is first grouped by $sourceIP, destinationIP, protocolService$ and the $sum$ aggregation operation is applied on the numerical features such as $responseBytes, requestBytes, duration ... $etc. This grouping for small intervals strikes a nice balance between unrealistically running a new inference for every change in the graph, i.e. every new connection in the network, and building a static graph for large intervals.

\begin{figure}
\centerline{\includegraphics[width=0.4\textwidth,height=0.12\textheight, trim={3.5cm 2.5cm 2.4cm 2.5cm},clip]{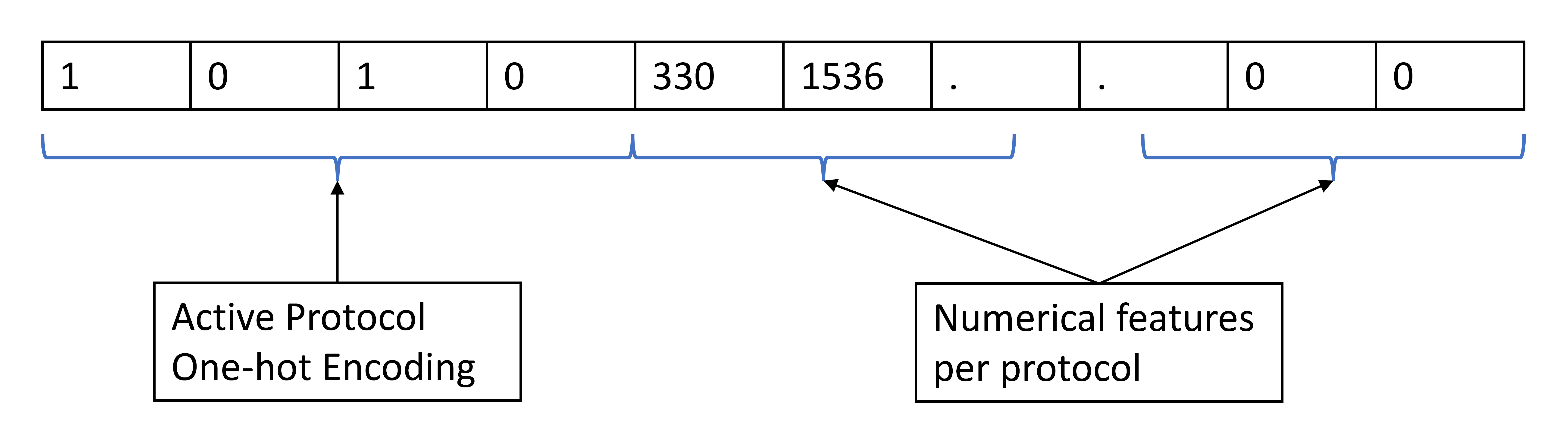}}
\caption{Edge input feature vector}
\label{fig:edgefeat}
\end{figure}

The structure of the graph is then as follows:
\begin{itemize}
    \item Node: each node represents an IP.
    \item Edge: There is at most one edge connection between two nodes if at least one connection happened between them using any protocol.
    \item Node Features: we do not have any node features in our data.
    \item Edge Features: each edge has a feature vector of the following format: the first part of the feature vector is a one-hot encoding of all the protocols seen between the source and destination nodes, and the rest is the concatenation of the aggregated numerical features for all protocols under study, as shown in Fig. \ref{fig:edgefeat}.
\end{itemize}

Since each node represents an IP, and in most deployments we do not have access to end-point data, there are no node features per se. This is a stark departure from existing models and datasets, where node features are the most important part of the data. Instead, in networking applications, almost all data is contained in the edge features, such as protocol information and traffic volume.

\subsubsection{GNN Input Layer with Edge Features Only}
\label{sssec: inputLayer}
Since in our graph the nodes are feature-less, we propose a new input layer that works solely on the edge features and can use them to generate preliminary node feature vectors.

\begin{figure}
\centerline{\includegraphics[width=0.4\textwidth,height=0.2\textheight, trim={3.2cm 3.1cm 4.9cm 3cm},clip]{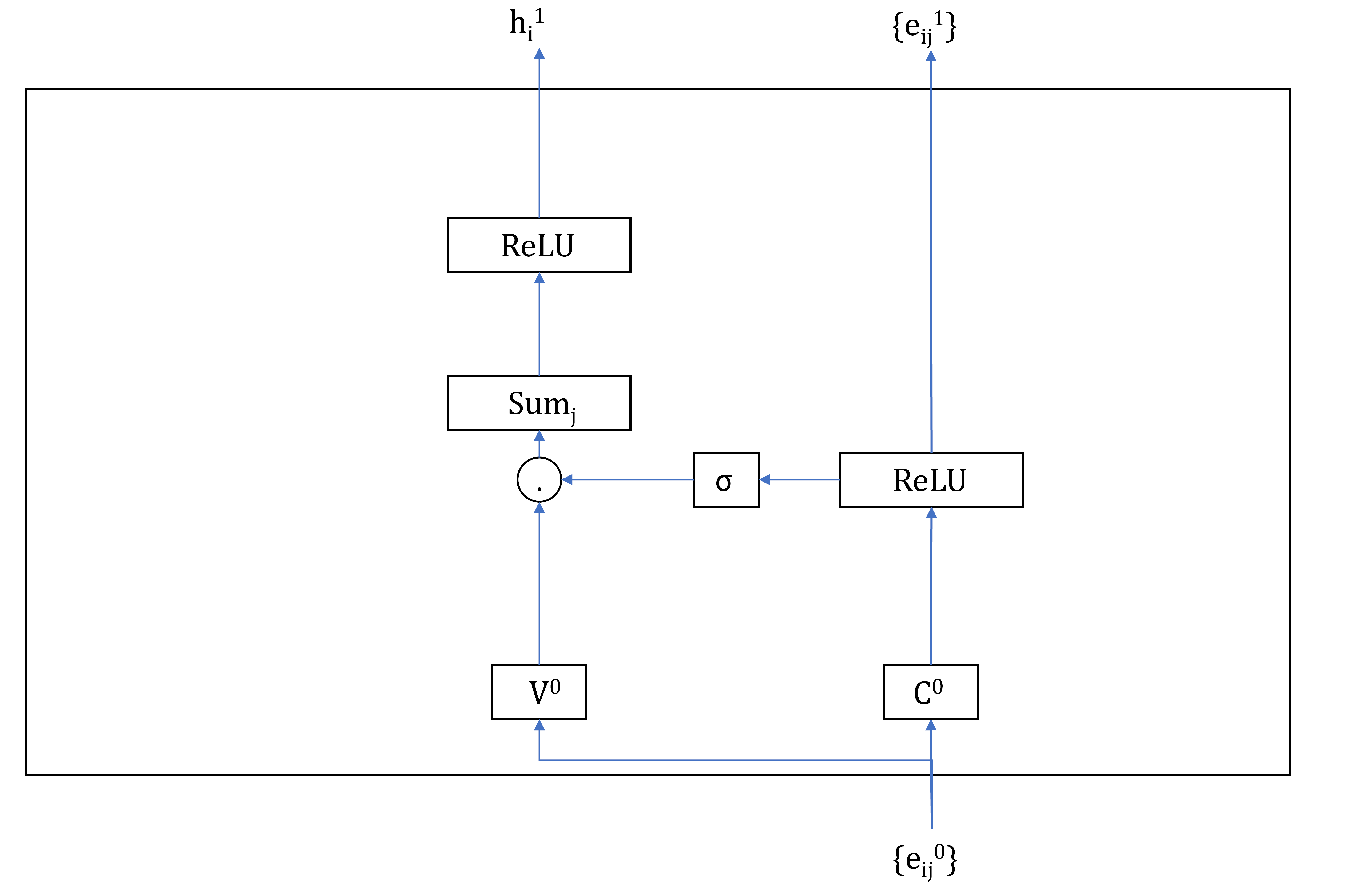}}
\caption{GNN Input Layer with Edge Features Only}
\label{fig:firstLayer}
\end{figure}

The input layer functions as follows:
\begin{equation}
    h_i^{1} = RELU\left(BN\left( \sum_{j \in \mathcal{N}_i} e_{ij}^1 \odot V^0 e_{ij}^0 \right) \right)
\end{equation}
where $W^0$ is a trainable weight matrix, $BN$ stands for batch normalization, $\odot$ is the Hadamard product, and the edge gates $e_{ij}^1$ are defined as 
\begin{equation}
\begin{aligned}
    & e_{ij}^1 = \frac{\sigma\left( \hat{e}_{ij}^1 \right)}
    {\sum_{j' \in \mathcal{N}_i} \sigma(\hat{e}_{ij'}^1) + \epsilon} \\
    & \hat{e}_{ij}^1 = e_{ij}^{0} + RELU\left(BN\left( C^0 e_{ij}^{0} \right) \right)
\end{aligned}
\end{equation}
and shown in Fig. \ref{fig:firstLayer}.

\subsubsection{GatedGCN Convolutional Layers}
\begin{figure}
\centerline{\includegraphics[width=0.4\textwidth,height=0.2\textheight, trim={5.3cm 3.1cm 4.9cm 3cm},clip]{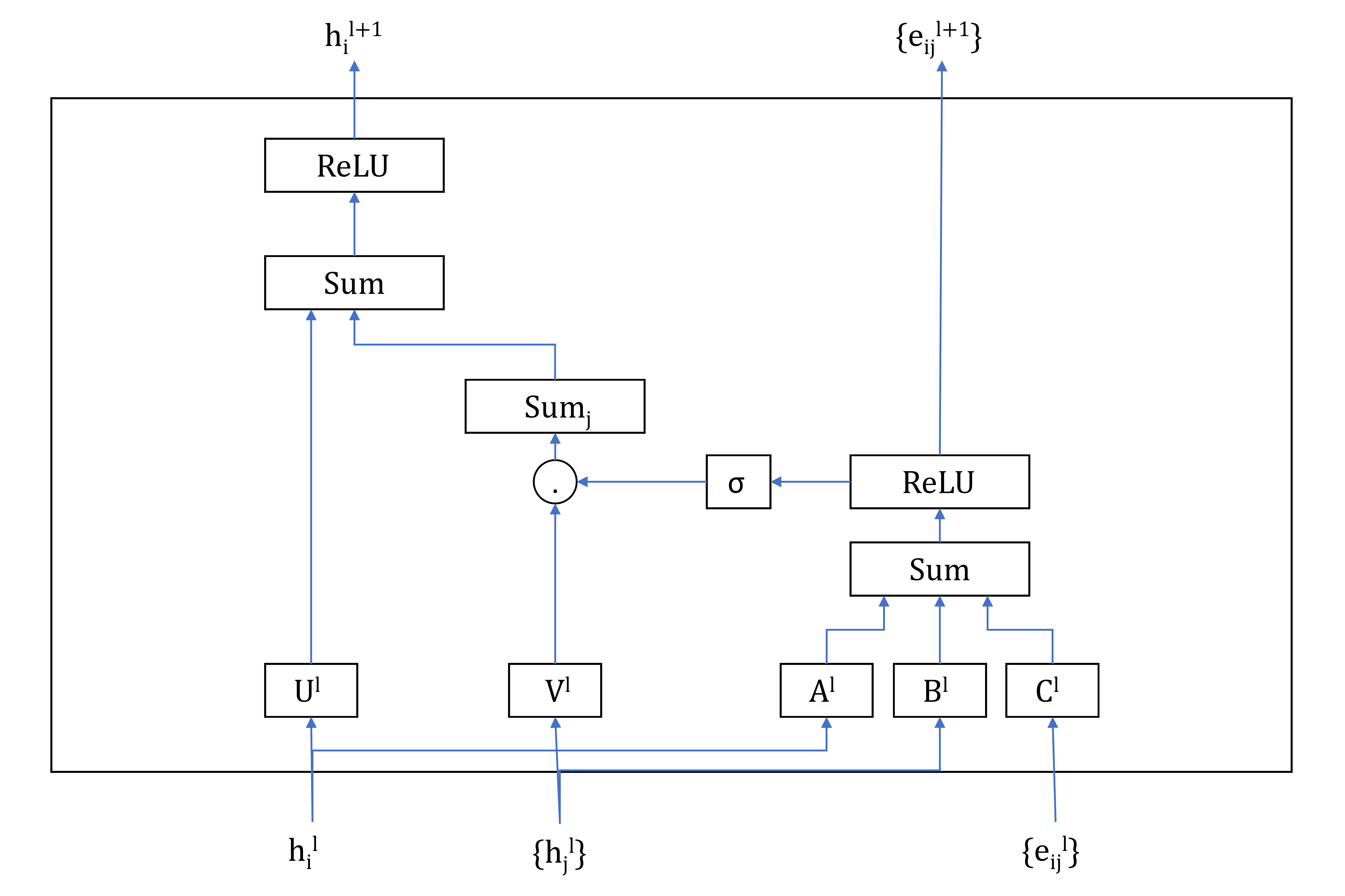}}
\caption{GatedGCN Convolutional Layer}
\label{fig:convLayer}
\end{figure}

Following the input layer, the other layers are the same as the convolutional layers in the original GatedGCN model in \cite{bresson2017residual}. These layers take the form 
\begin{equation}
    h_i^{l+1} = h_i^l + RELU\left(BN\left(U^l h_i^l 
    + \sum_{j \in \mathcal{N}_i} e_{ij}^l \odot V^l h_j^l \right) \right)
\end{equation}
where $U^l, V^l$ are trainable weight matrices, and the edge gates $e_{ij}^l$ are defined as 
\begin{equation}
\begin{aligned}
    & e_{ij}^l = \frac{\sigma\left( \hat{e}_{ij}^l \right)}
    {\sum_{j' \in \mathcal{N}_i} \sigma(\hat{e}_{ij'}^l) + \epsilon} \\
    & \hat{e}_{ij}^l = \hat{e}_{ij}^{l-1} + RELU\left(BN\left(A^l h_i^{l-1} + B^l h_j^{l-1} + C^l \hat{e}_{ij}^{l-1} \right) \right)
\end{aligned}
\end{equation}

and shown in Fig. \ref{fig:convLayer}. In both types of layers, the edge gates $e_{ij}^l$ can be seen as a soft-attention mechanism \cite{dwivedi2020benchmarking}.

\subsubsection{Loss Function}
\begin{figure}
\centerline{\includegraphics[width=0.5\textwidth,height=0.175\textheight, trim={0.2cm 1.1cm 1.8cm 1cm},clip]{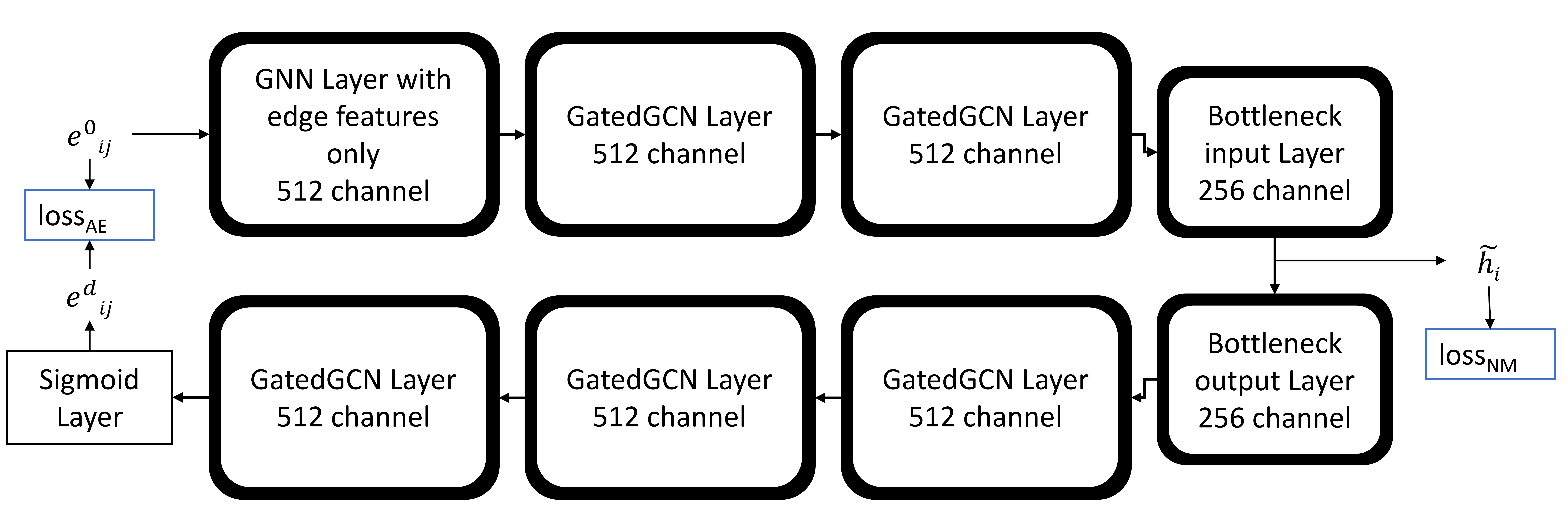}}
\caption{GNN Model}
\label{fig:model}
\end{figure}

Our loss function is composed of two parts:
\begin{itemize}
    \item Edge Feature Reconstruction: this is main part of the loss function and represents the auto-encoder learning objective. The goal here is to reconstruct the edge features from the low dimension representation as follows:
    \begin{equation}
        loss_{AE} = \lambda_{AE} BinaryCrossEntropy\left(e_{ij}^0, e_{ij}^d\right)
    \end{equation}
    where $e_{ij}^0$ are the input edge features and $e_{ij}^d$ are the feature representations from the last layer of the decoder.
    
    \item Neighborhood Embedding Match: similar to other graph auto-encoder approaches \cite{salehi2019graph}, we would like embeddings of neighboring nodes to be similar:
    \begin{equation}
        loss_{NM} = - \lambda_{NM} \sum_{i=1}^N \sum_{j \in \mathcal{N}_i} 
        \log \left( \frac{1}{1 + exp(- \mathbf{\Tilde{h}}_i^T \mathbf{\Tilde{h}}_j)} \right)
    \end{equation}
    where $\Tilde{h}_i, \Tilde{h}_j$ are the embeddings of nodes $i, j$ respectively.
    
    
\end{itemize}
The total loss is simply the sum of all the loss functions:
\begin{equation}
    loss = loss_{AE} + loss_{NM} 
\end{equation}
The complete GNN model is shown in Fig. \ref{fig:model}. The sparsity of the input edge feature vectors proved to be a consistent challenge for the stability and convergence of the GNN learning process. To handle this issue, we have found it best to normalize the input features, use a $Sigmoid$ layer as the output of the decoder and and $BinaryCrossEntropy(.,.)$ as the loss function between the two.

\section{Results}
\label{sec:results}
In this paper, we use a private dataset from a medium-sized enterprise network. This network has around 50 users using  primarily windows machines, and a few internal servers.  The traffic sensor is located such that it can capture both the traffic between internal hosts and between internal and external hosts, so-called east-west traffic and north-south traffic.

We conduct two main experiments to judge the ability of our model to measure similarity between new and previously observed nodes. In particular, given a set of IPs that we know to belong to machines of similar functionality, we remove a subset during the training phase and only introduce them back during the testing phase. We do so by removing all conn logs where either the $sourceIP$ or the $destinationIP$ is the part of the subset to be removed. We then expect the model to produce high similarity between the whole set of such IPs, not only the ones seen during training. We conduct this experiment using two sets of IPs:
\begin{itemize}
    \item Local DNS servers: the network under study has two local DNS servers, 192.168.2.15 and 192.168.2.19. We remove 192.168.2.19 during training.
    
    \item Root DNS servers: these servers are the highest in the hierarchy of the internet DNS servers\footnote{\url{https://en.wikipedia.org/wiki/Root_name_server#Root_server_addresses}}. These IPs are $198.41.0.4, 199.9.14.201, 192.33.4.12, \\ 
    199.7.91.13, 192.203.230.10, 192.5.5.241, \\
    192.112.36.4, 198.97.190.53, 192.36.148.17,\\
    192.58.128.30, 193.0.14.129, 199.7.83.42, 202.12.27.33$. We remove the last four IPs during training time and measure their similarity with the first nine IPs during test time.
\end{itemize}
In all our experiments we adopt the cosine similarity measure defined as
\begin{equation}
    d(\Tilde{h}_i, \Tilde{h}_j) = \frac{\Tilde{h}_i^T \Tilde{h}_j}{||\Tilde{h}_i||.||\Tilde{h}_j ||}
\end{equation}
We also conduct all experiments using both $loss_{AE}$ and $loss_{AE} + loss_{NM}$ as the loss functions, with $\lambda_{AE}=1.0, \lambda_{NM}=0.01$.

In Table \ref{tab:simil1519}, we show the similarity between the two IPs, 192.168.2.15 and 192.168.2.19 on the test graphs. Except for the first row, the similarity between these two IPs quickly reaches very high values, indicating the ability of the model to recognize the similar behavior even though only one IP was seen before. We posit that for shorted periods, e.g. 10 minutes, the amount of traffic seen might not be representative enough of the typical behavior of the machine and hence one model produced low similarity between the two IPs in this case. 

In Table \ref{tab:similroot} we show a similar set of results but for the root DNS servers. Similarity is averaged across all possible pairs when the first IP is one of the 9 root IPs seen during both training and testing while the second IP belongs to the set of 4 root IPs seen only during testing. The same observation holds here that the model was able to recognize the similarity between these public IPs due to their similar functionality. We also see high similarity here even for the shorter period of 10 minutes. We explain that as public DNS servers would only receive one kind of traffic, that is DNS queries. However, we have more visibility into the traffic between the local machines, including local DNS servers, and hence it takes a longer interval to see the dominant traffic pattern for local machines. We would like to note that although we have a relatively high number of test graphs, the inference passes are independent across each graph. In other words, the model does not need to see a large number of graphs including the IP(s) under study, in fact, once the model is trained, it can generate a high-quality embedding for an IP from a single test graph. 

\begin{figure}[!htb]
\centering
\includegraphics[width=0.45\textwidth,height=0.25\textheight, trim={1.3cm 1.1cm 3cm 2cm},clip]{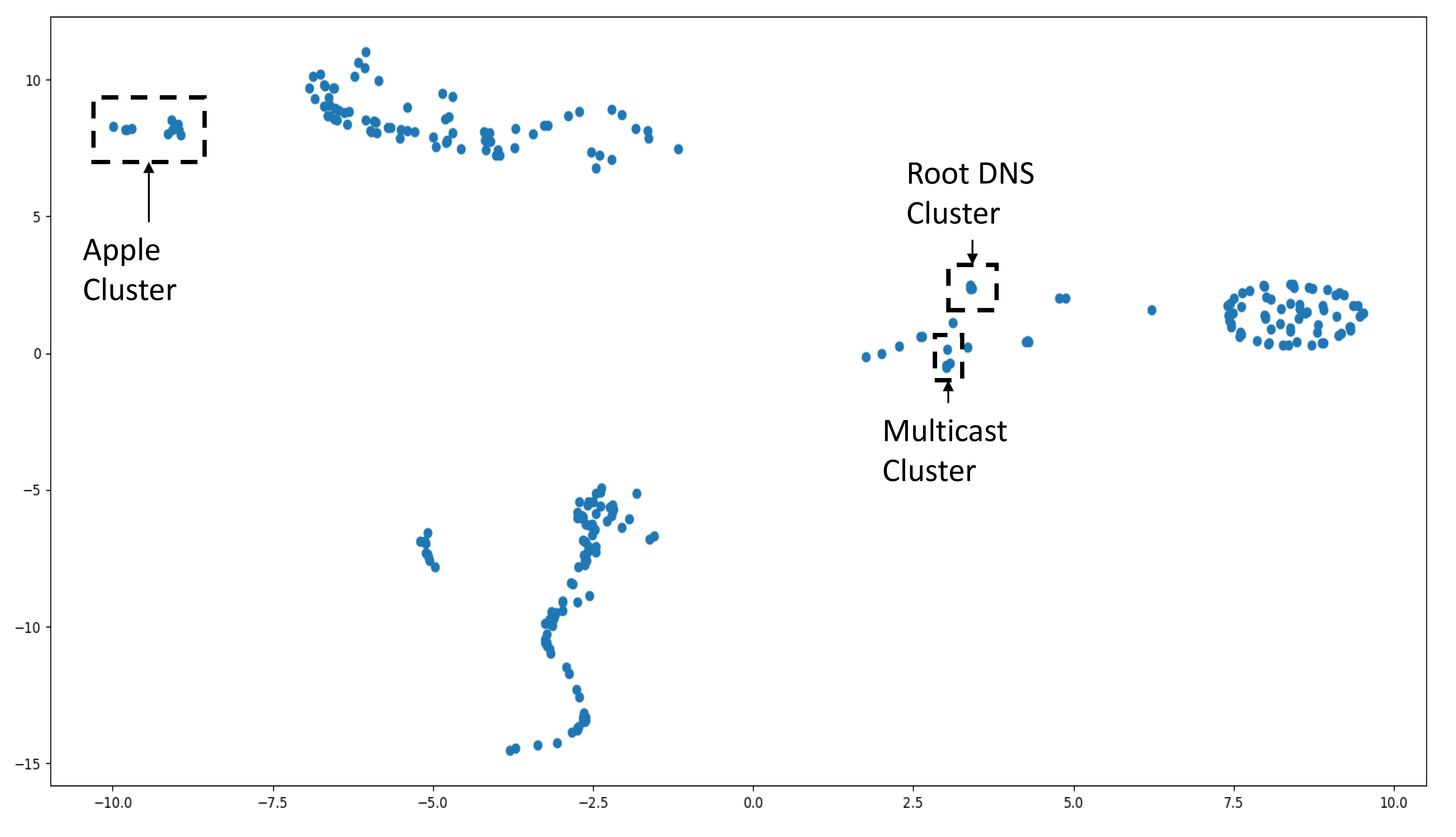}
\caption{t-SNE visualization of all IPs with highlighted clusters}
\label{fig:all_visual}
\end{figure}

\begin{figure}[!htb]
\centering
\includegraphics[width=0.5\textwidth,height=0.2\textheight, trim={1.3cm 1.1cm 12cm 2cm},clip]{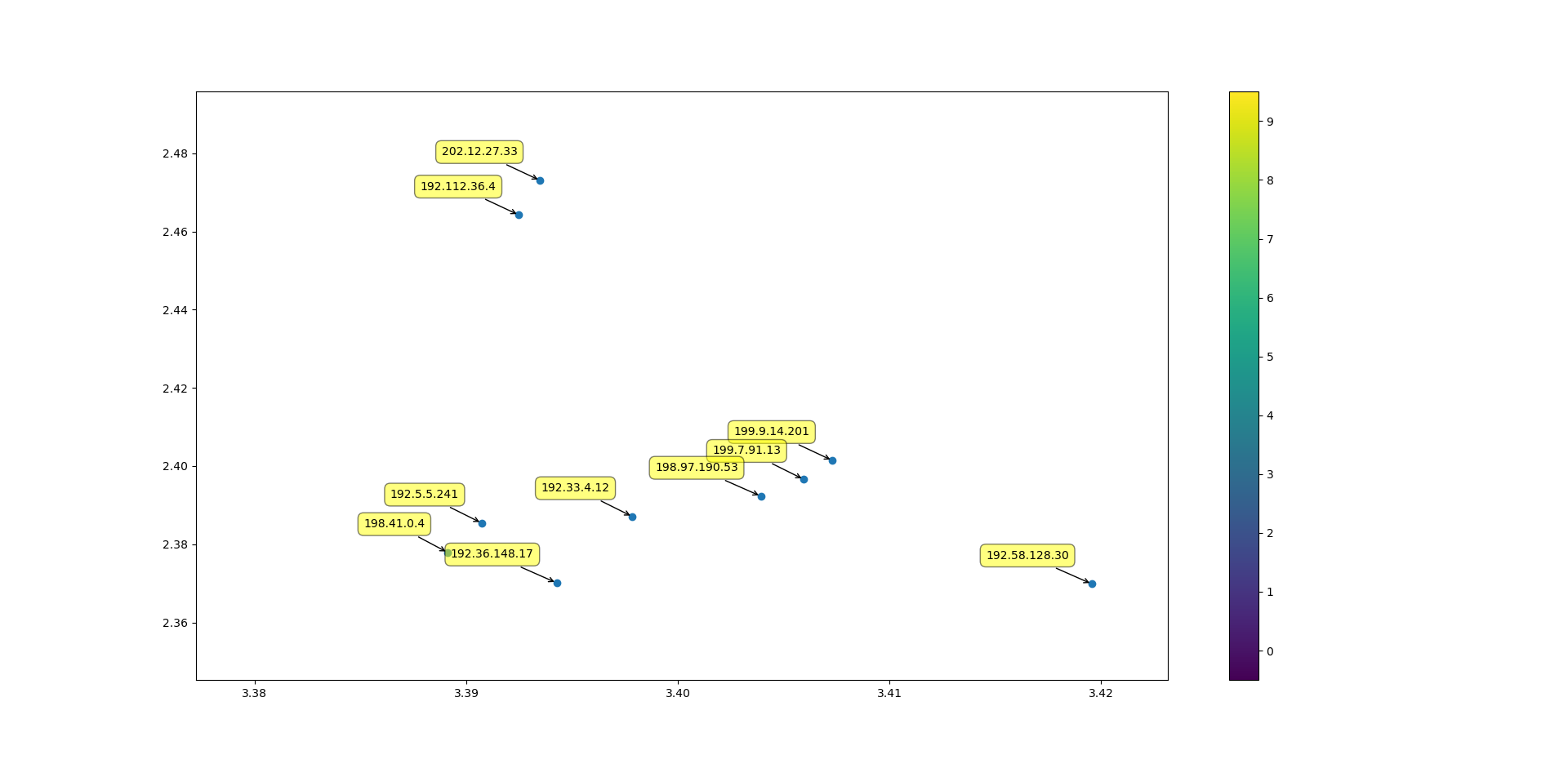}
\caption{t-SNE visualization of Root DNS Cluster}
\label{fig:root_ips}
\end{figure}

\begin{figure}[!htb]
\centering
\includegraphics[width=0.5\textwidth,height=0.2\textheight, trim={1.3cm 1.1cm 12cm 2cm},clip]{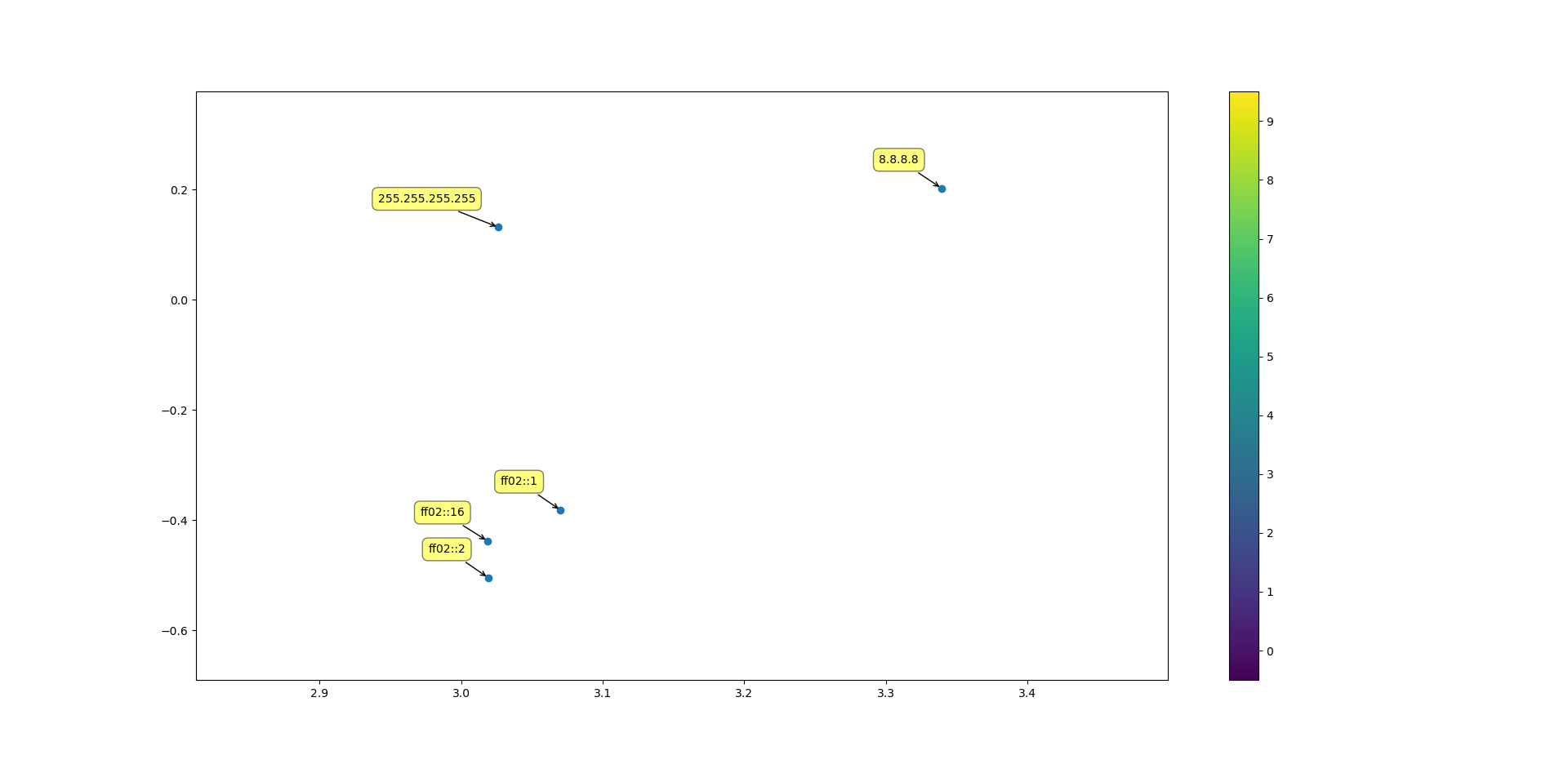}
\caption{t-SNE visualization of Broadcast and Multicast Cluster}
\label{fig:mcast_ips}
\end{figure}

\begin{figure}[!htb]
\centering
\includegraphics[width=0.5\textwidth,height=0.2\textheight, trim={1.3cm 1.1cm 12cm 2cm},clip]{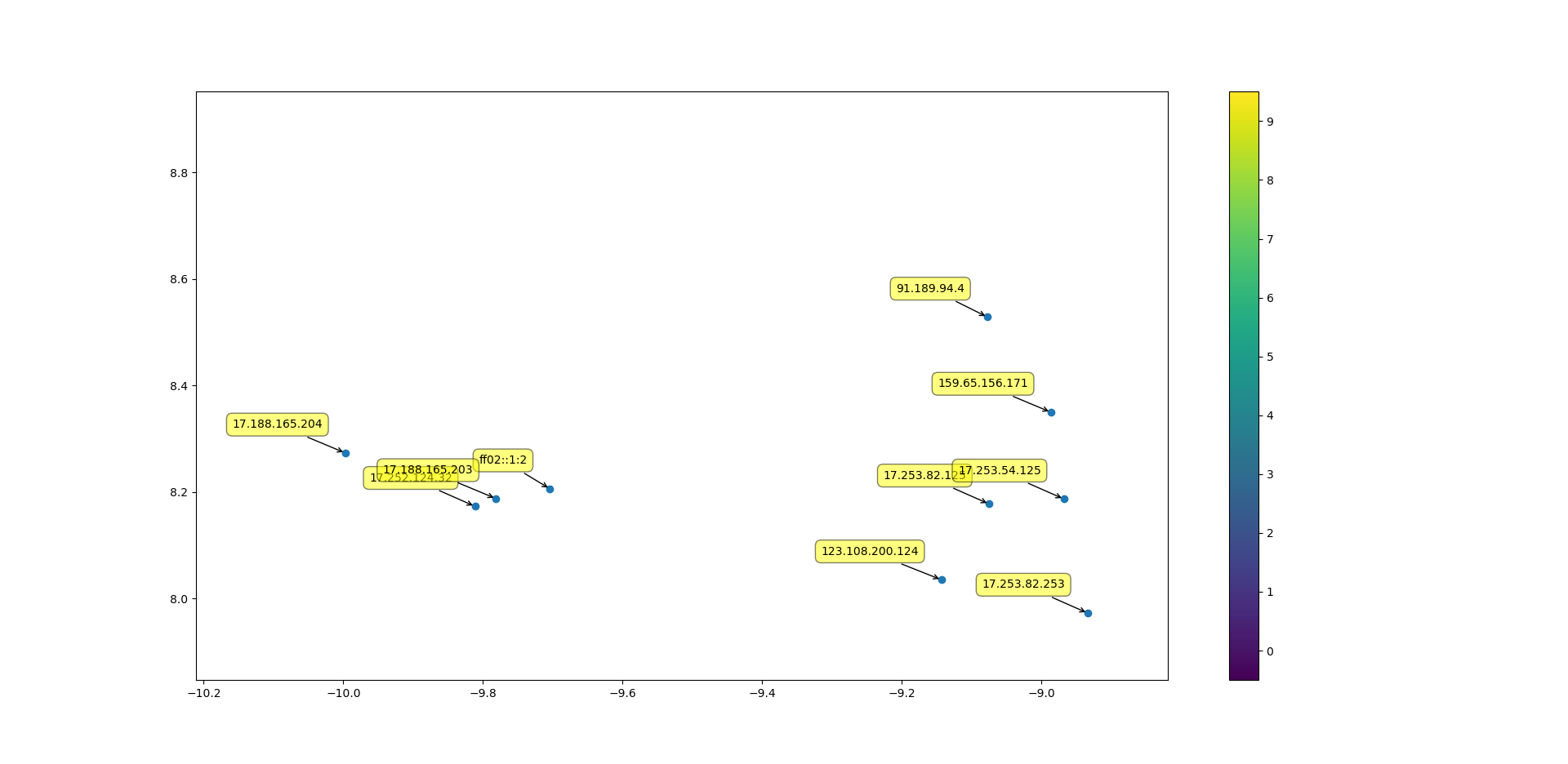}
\caption{t-SNE visualization of Apple Cluster}
\label{fig:17_ips}
\end{figure}

We also provide visualizations of some interesting clusters of similar IPs. Fig. \ref{fig:all_visual} shows the visualizations of all IPs in the network during a single interval with some interesting clusters highlighted, these clusters are zoomed on in the figures to follow. Fig. \ref{fig:root_ips} has a cluster including the root DNS IPs discussed before. Fig. \ref{fig:mcast_ips} has the broadcast IP 255.255.255.255, and all the other IPv6 addresses are multicast IPs. In Fig. \ref{fig:17_ips}, all the IPs of the form 17.* belong to Apple. In all four figures, t-SNE \cite{maaten2008visualizing} is used for the 2-dimensional visualization.

\begin{table*}[hbt]
  \begin{center}
    \caption{Similarity between two local DNS Servers when only one is seen during training}
    \label{tab:simil1519}
    \begin{tabular}{L{3cm} | L{4cm} | L{4cm} | L{3cm}}
      \toprule 
      \textbf{Graph Interval Length} & \textbf{$loss_{AE}$ Similarity} & \textbf{$loss_{AE} + loss_{NM}$ Similarity} & \textbf{Number of Test Graphs}\\
      \toprule 
      10 minutes & 0.7060738 $\pm$ 0.34426507  & 0.9862463 $\pm$ 0.01696037 & 207 \\
      \midrule 
      30 minutes & 0.9743041 $\pm$ 0.04809975 & 0.9869821 $\pm$ 0.01701941 & 70 \\
      \midrule 
      60 minutes & 0.9647852 $\pm$ 0.06598776 & 0.9906003 $\pm$ 0.01601414 & 35 \\
      \midrule 
      120 minutes & 0.9880788 $\pm$ 0.030548556 & 0.9803449 $\pm$ 0.05673271 & 18 \\
      \bottomrule 
    \end{tabular}
  \end{center}
\end{table*}

\begin{table*}[hbt]
  \begin{center}
    \caption{Average Similarity between root DNS Servers }
    \label{tab:similroot}
    \begin{tabular}{L{3cm} | L{4cm} | L{4cm} | L{3cm}}
      \toprule 
      \textbf{Graph Interval Length} & \textbf{$loss_{AE}$ Similarity} & \textbf{$loss_{AE} + loss_{NM}$ Similarity} & \textbf{Number of Test Graphs}\\
      \toprule 
      10 minutes & 0.9654844 $\pm$ 0.04811225 & 0.9918937 $\pm$ 0.00860384 & 207 \\
      \midrule 
      30 minutes &  0.9791588 $\pm$ 0.03466465 & 0.9879824 $\pm$ 0.01978309 & 70 \\
      \midrule 
      60 minutes & 0.9768600 $\pm$ 0.04310611 & 0.9870570 $\pm$ 0.01964757 & 35 \\
      \midrule 
      120 minutes & 0.9930404 $\pm$ 0.01426881 & 0.9593274 $\pm$ 0.09711087 & 18 \\
      \bottomrule 
    \end{tabular}
  \end{center}
\end{table*}

\section{Conclusion and Future Work}
\label{sec:conclusion}
In this paper, we introduced the first study on applying GNNs for the embedding of network assets, represented as IPs. We identified a few challenges that render this problem uniquely different from both NLP-based network embedding and the more common use-cases of GNNs. These challenges revolve mainly around the inductive nature of a model being significantly more important compared to text data, and the lack of node features in typical deployments of such a model. We then introduced a new input layer to handle our data, and the necessary output layer and loss function to stabilize the training process. The outcome is an efficient embedding for each asset in the network incorporating its own as well as its neighbors traffic behavior. Our results show that the model is able to properly represent similar IPs even though only a subset of those IPs has been seen before during the training phase. The model is also able to highlight interesting IP clusters, such DNS servers, broadcast/multicast IPs and Apple IPs.

An extension to our approach is to introduce node features in the system. These node features will have to come from end-point data, such as Windows event log. The model can then learn better embeddings representing both the end-point and the network behavior of each asset in the network.

\bibliographystyle{IEEEtrannames}
\bibliography{references}

\begin{thebibliography}{10}
\providecommand{\url}[1]{#1}
\csname url@rmstyle\endcsname
\providecommand{\newblock}{\relax}
\providecommand{\bibinfo}[2]{#2}
\providecommand\BIBentrySTDinterwordspacing{\spaceskip=0pt\relax}
\providecommand\BIBentryALTinterwordstretchfactor{4}
\providecommand\BIBentryALTinterwordspacing{\spaceskip=\fontdimen2\font plus
\BIBentryALTinterwordstretchfactor\fontdimen3\font minus
  \fontdimen4\font\relax}
\providecommand\BIBforeignlanguage[2]{{%
\expandafter\ifx\csname l@#1\endcsname\relax
\typeout{** WARNING: IEEEtran.bst: No hyphenation pattern has been}%
\typeout{** loaded for the language `#1'. Using the pattern for}%
\typeout{** the default language instead.}%
\else
\language=\csname l@#1\endcsname
\fi
#2}}

\bibitem{Ghafir2014}
I.~Ghafir and V.~Prenosil, ``Advanced persistent threat attack detection: an
  overview,'' \emph{International Journal of Advances in Computer Networks and
  Its Security (IJCNS)}, vol.~4, no.~4, p. 5054, 2014.

\bibitem{cao2015preemptive}
P.~Cao, E.~Badger, Z.~Kalbarczyk, R.~Iyer, and A.~Slagell, ``Preemptive
  intrusion detection: Theoretical framework and real-world measurements,'' in
  \emph{Proceedings of the 2015 Symposium and Bootcamp on the Science of
  Security}, 2015, pp. 1--12.

\bibitem{sommer2010outside}
R.~Sommer and V.~Paxson, ``Outside the closed world: On using machine learning
  for network intrusion detection,'' in \emph{IEEE symposium on security and
  privacy}, 2010, pp. 305--316.

\bibitem{veeramachaneni2016ai}
K.~Veeramachaneni, I.~Arnaldo, V.~Korrapati, C.~Bassias, and K.~Li, ``Ai\^{} 2:
  training a big data machine to defend,'' in \emph{2016 IEEE 2nd International
  Conference on Big Data Security on Cloud (BigDataSecurity)}.\hskip 1em plus
  0.5em minus 0.4em\relax IEEE, 2016, pp. 49--54.

\bibitem{shashanka2016user}
M.~Shashanka, M.-Y. Shen, and J.~Wang, ``User and entity behavior analytics for
  enterprise security,'' in \emph{2016 IEEE International Conference on Big
  Data (Big Data)}.\hskip 1em plus 0.5em minus 0.4em\relax IEEE, 2016, pp.
  1867--1874.

\bibitem{julisch2003clustering}
K.~Julisch, ``Clustering intrusion detection alarms to support root cause
  analysis,'' \emph{ACM transactions on information and system security
  (TISSEC)}, vol.~6, no.~4, pp. 443--471, 2003.

\bibitem{coull2011measuring}
S.~E. Coull, F.~Monrose, and M.~Bailey, ``On measuring the similarity of
  network hosts: Pitfalls, new metrics, and empirical analyses.'' in
  \emph{NDSS}, 2011.

\bibitem{shittu2016mining}
R.~O. Shittu, ``Mining intrusion detection alert logs to minimise false
  positives \& gain attack insight,'' Ph.D. dissertation, City University
  London, 2016.

\bibitem{ring2017ip2vec}
M.~Ring, A.~Dallmann, D.~Landes, and A.~Hotho, ``Ip2vec: Learning similarities
  between ip addresses,'' in \emph{2017 IEEE International Conference on Data
  Mining Workshops (ICDMW)}.\hskip 1em plus 0.5em minus 0.4em\relax IEEE, 2017,
  pp. 657--666.

\bibitem{goldberg2014word2vec}
Y.~Goldberg and O.~Levy, ``word2vec explained: deriving mikolov et al.'s
  negative-sampling word-embedding method,'' \emph{arXiv preprint
  arXiv:1402.3722}, 2014.

\bibitem{burr2020detection}
B.~Burr, S.~Wang, G.~Salmon, and H.~Soliman, ``On the detection of persistent
  attacks using alert graphs and event feature embeddings,'' in \emph{IEEE/IFIP
  NOMS}, 2020, pp. 1--4.

\bibitem{bazzi2002modelling}
I.~Bazzi, ``Modelling out-of-vocabulary words for robust speech recognition,''
  Ph.D. dissertation, Massachusetts Institute of Technology, 2002.

\bibitem{kim2016character}
Y.~Kim, Y.~Jernite, D.~Sontag, and A.~M. Rush, ``Character-aware neural
  language models,'' in \emph{Proceedings of the Thirtieth AAAI Conference on
  Artificial Intelligence}, 2016, pp. 2741--2749.

\bibitem{dwivedi2020benchmarking}
V.~P. Dwivedi, C.~K. Joshi, T.~Laurent, Y.~Bengio, and X.~Bresson,
  ``Benchmarking graph neural networks,'' \emph{arXiv preprint
  arXiv:2003.00982}, 2020.

\bibitem{wu2020comprehensive}
Z.~Wu, S.~Pan, F.~Chen, G.~Long, C.~Zhang, and S.~Y. Philip, ``A comprehensive
  survey on graph neural networks,'' \emph{IEEE Transactions on Neural Networks
  and Learning Systems}, 2020.

\bibitem{lecun2015deep}
Y.~LeCun, Y.~Bengio, and G.~Hinton, ``Deep learning,'' \emph{nature}, vol. 521,
  no. 7553, pp. 436--444, 2015.

\bibitem{hamilton2017representation}
W.~L. Hamilton, R.~Ying, and J.~Leskovec, ``Representation learning on graphs:
  Methods and applications,'' \emph{arXiv preprint arXiv:1709.05584}, 2017.

\bibitem{hamilton2017inductive}
W.~Hamilton, Z.~Ying, and J.~Leskovec, ``Inductive representation learning on
  large graphs,'' in \emph{NIPS}, 2017, pp. 1024--1034.

\bibitem{bresson2017residual}
X.~Bresson and T.~Laurent, ``Residual gated graph convnets,'' \emph{arXiv
  preprint arXiv:1711.07553}, 2017.

\bibitem{salehi2019graph}
A.~Salehi and H.~Davulcu, ``Graph attention auto-encoders,'' \emph{arXiv
  preprint arXiv:1905.10715}, 2019.

\bibitem{maaten2008visualizing}
L.~v.~d. Maaten and G.~Hinton, ``Visualizing data using t-sne,'' \emph{Journal
  of machine learning research}, vol.~9, no. Nov, pp. 2579--2605, 2008.

\end{thebibliography}

\end{document}